\documentclass{article}

\usepackage[main, final]{neurips_2026}

\usepackage[utf8]{inputenc} 
\usepackage[T1]{fontenc}    
\usepackage{hyperref}       
\usepackage{url}            
\usepackage{booktabs}       
\usepackage{amsfonts}       
\usepackage{nicefrac}       
\usepackage{microtype}      
\usepackage[dvipsnames]{xcolor}         

\usepackage{graphicx}
\usepackage{booktabs}
\usepackage{comment}
\usepackage{bm}
\usepackage{makecell}
\usepackage{multirow}
\usepackage{xspace}
\usepackage[accsupp]{axessibility}  
\usepackage{graphicx}
\usepackage{subcaption}

\makeatletter
\DeclareRobustCommand\onedot{\futurelet\@let@token\@onedot}
\def\@onedot{\ifx\@let@token.\else.\null\fi\xspace}

\def\eg{\emph{e.g}\onedot} \def\Eg{\emph{E.g}\onedot}
\def\ie{\emph{i.e}\onedot} 
\def\cf{\emph{c.f}\onedot}

\makeatother

\usepackage{array}
\newcolumntype{C}[1]{>{\centering\arraybackslash}m{#1}}
\newcolumntype{M}[1]{>{\centering\arraybackslash}p{#1}}

\definecolor{darkgreen}{RGB}{0, 100, 0}   
\definecolor{DarkMagenta}{rgb}{0.7, 0.0, 0.7}

\newif\ifshowedits
\newcommand{\addeditor}[3]{%
  \definecolor{#1color}{rgb}{#3}
  \expandafter\newcommand\csname #1\endcsname[1]{%
  \ifshowedits
    {\color{#1color} ##1}%
  \else
    {##1}%
  \fi
  }%
  \expandafter\newcommand\csname #1rmk\endcsname[1]{%
  \ifshowedits
    {\color{#1color} {\bf [#2: ##1]}}
  \fi
  }%
  \expandafter\newcommand\csname #1rpl\endcsname[2]{%
  \ifshowedits
    {{\color{#1color} ##1} \sout{##2}}
  \else
    {##1}
  \fi
  }%
}

\title{Failing Forward: Adaptive Failure-Informed Learning for Vision-Language-Action Models}

\author{
Meng Zheng$^{1}$,
Samhita Marri$^{1,2}$*,
Anwesa Choudhuri$^{1}$,
Benjamin Planche$^{1}$,
Zhongpai Gao$^{1}$, \\
\textbf{Van Nguyen Nguyen}$^{1}$, 
\textbf{Terrence Chen}$^{1}$,
\textbf{Girish Chowdhary}$^{2}$ and
\textbf{Ziyan Wu}$^{1}$\\[0.3em]
$^{1}$United Imaging Intelligence, Boston, MA, USA \\
$^{2}$University of Illinois Urbana-Champaign, Urbana, IL, USA\\
}

\begin{document}

\begingroup
\renewcommand{\thefootnote}{\fnsymbol{footnote}}
\footnotetext[1]{This work was carried out during the internship of Samhita Marri at United Imaging Intelligence, Boston, MA}
\endgroup
\setcounter{footnote}{0}

\addeditor{zhongpai}{ZP}{0.7, 0.0, 0.7}
\addeditor{ziyan}{ZW}{0.0, 0.5, 0.0}
\addeditor{benjamin}{BP}{0.8, 0.4, 0.1}
\addeditor{meng}{MZ}{0.5, 0.4, 0.1}
\addeditor{anwesa}{AC}{0.3, 0.8, 0.1}
\addeditor{nguyen}{VN}{0.9, 0.1, 0.9}
\showeditstrue

\maketitle

\begin{abstract}
  Vision-language-action (VLA) models provide a promising paradigm for scalable robotic manipulation, yet their reliance on success-only behavioral cloning leaves them brittle; lacking corrective training signals, minor execution errors rapidly compound into unrecoverable, out-of-distribution failures. To address this limitation, we propose Adaptive Failure-Informed Learning (AFIL), an end-to-end framework that leverages failure trajectories as adaptive negative guidance for diffusion- and flow-based VLA policies. AFIL uses a pretrained VLA to generate failure rollouts online, avoiding the need for handcrafted failure-mode design or human-in-the-loop recovery. It then jointly trains Dual Action Generators (DAGs) for successful and failed behaviors while sharing a common vision-language backbone, enabling efficient failure-aware policy learning with limited parameter overhead. 
During sampling, the failure generator adaptively steers action generation away from failure-prone regions and toward more reliable success modes, with guidance strength determined by the per-diffusion-step distance between success and failure distributions. Experiments across in-domain and out-of-domain robotic manipulation tasks, covering both short- and long-horizon settings, show that AFIL consistently improves task success rates and robustness over existing VLA baselines, demonstrating its effectiveness, efficiency, and generality.
\end{abstract}

\section{Introduction}
\label{sec:intro}
Robotic manipulation has long been a central challenge in robotics and artificial intelligence, requiring agents to perceive complex environments, reason over long horizons, and execute precise control under uncertainty. Conventional learning-based manipulation policies, including behavior cloning and reinforcement learning (RL), have achieved notable success in structured settings, but often require task-specific reward design, extensive interaction data, and careful environment engineering, limiting their scalability and generalization to diverse real-world tasks. Recently, vision-language-action (VLA) models have attracted growing attention in robotics \citep{pertsch2025fast,black2024pi_0,intelligence2025pi,liu2025rdtb,chi2025visuomotor,zitkovich2023rt,zhao2025cot}, particularly following the release of OpenVLA \citep{kim2024openvla}, which demonstrated the feasibility of leveraging large-scale pretrained vision-language representations for embodied decision-making. 

\begin{figure*}[!t]
    \centering
    \includegraphics[width=\textwidth]{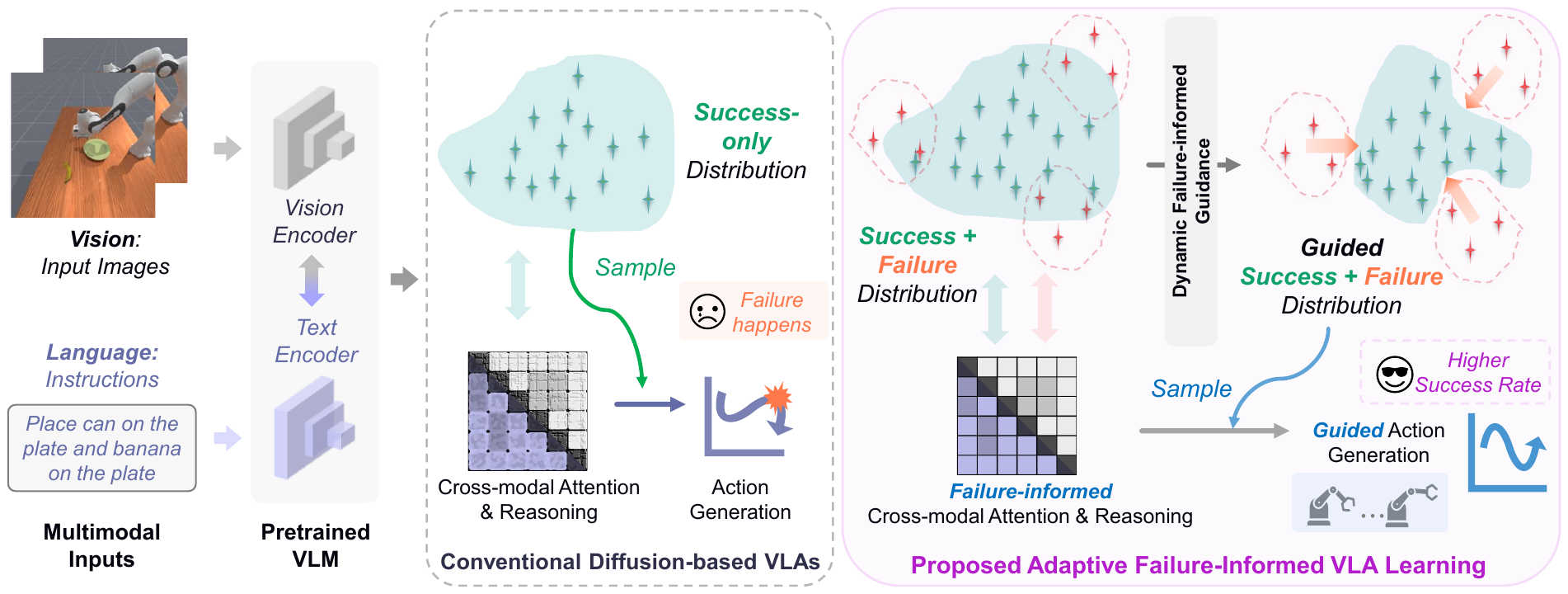}
    \caption{Overview of the proposed Adaptive Failure-Informed Learning (AFIL) pipeline. AFIL uses online-generated failure trajectories as adaptive negative guidance, steering action generation away from failure-prone regions and toward more reliable success modes.}
    \label{fig:teaser}
\end{figure*}

Despite their promise, VLA models often struggle to generalize during deployment, as training data typically consists primarily of teleoperated, success-only trajectories collected in confined environments \citep{kim2024openvla,o2024open,james2020rlbench,mandlekar2021robomimic}. Just as a smooth sea never made a skilled sailor, \textit{a smooth demonstration never made a skilled robot}. Such data can cause policies to overfit to narrow behavioral manifolds and lack corrective signals for recovering from suboptimal or unseen states, resulting in brittle execution at inference time. Recent work has therefore highlighted the value of learning from failure data \citep{lu2025robofac,lin2025failsafe,huang2025failprogress,duan2024aha}, showing that unsuccessful trajectories can improve robustness and task success rates. However, existing approaches often rely on handcrafted or predefined failure modes; for example, \cite{duan2024aha} manually specifies categories such as incomplete grasp, slip, and translation failures for data generation. These designs require domain expertise and may fail to capture the diversity of realistic deployment-time failures. Moreover, recovery often depends on task-specific heuristics or real-time human intervention \citep{lin2025failsafe,duan2024aha,shi2024yell,zhang2024don}, limiting scalability and generalization to unseen tasks or embodiments.

Most VLA policies adopt diffusion-based \citep{DDPM_NeurIPS20,chi2025visuomotor} or flow-based \citep{lipman2023flow} generative architectures \citep{kim2024openvla,black2024pi_0,intelligence2025pi,chi2025visuomotor}, owing to their ability to model multimodal action distributions and capture complex dependencies between perception and control. 
However, since most existing diffusion-/flow-based VLA models are trained only on successful demonstrations, sampling from the learned success-conditioned distribution may be insufficient for robust deployment, as the policy lacks explicit signals to avoid nearby failure-prone actions. In image generation, guided diffusion methods such as classifier guidance (CG) \citep{CG_NeurIPS21} and classifier-free guidance (CFG) \citep{ho2021classifierfree,Rombach2021HighResolutionIS,chung2025cfg,Fu_2025_ICCV} improve conditional generation by steering samples toward desired semantic attributes. Negative prompting further extends this idea by repelling samples away from undesirable regions of the generation space \citep{gandikota2023erasing,schramowski2022safe,koulischer2025dynamic}. This naturally motivates an analogous strategy for VLA learning: treating failure trajectories as negative guidance to encourage the policy to avoid action distributions associated with unsuccessful outcomes.
Directly applying negative guidance to VLAs, however, is non-trivial \citep{pearce2023imitating,reuss2023goal}. Unlike image generation, robotic action generation is sequential and must produce temporally consistent actions conditioned on evolving observations. In addition, conventional negative guidance methods often rely on separate models for desired and undesired distributions, which is computationally expensive for VLA systems built on large vision-language model (VLM) backbones. These challenges motivate an efficient, unified failure-informed guidance mechanism tailored to diffusion- and flow-based VLA policies.

To address these challenges, we propose Adaptive Failure-Informed Learning (AFIL), an end-to-end framework for diffusion- and flow-matching VLA models that dynamically guides action generation using failure examples (\cf Figure~\ref{fig:teaser}). AFIL learns from mixed success and failure data by jointly training Dual Action Generator (DAG)-VLA for successful and failed behaviors while sharing a VLM backbone for visual perception and semantic grounding. 
During sampling, AFIL iteratively queries both success and failure action generators and uses the predicted failure actions as adaptive negative guidance, steering the generative process away from undesirable regions of the action distribution. Extensive experiments on both in-domain and out-of-domain robotic manipulation tasks show that AFIL consistently improves task success rates and robustness over strong VLA baselines, while introducing only moderate additional parameters. Our contributions are summarized as follows:
\begin{itemize}
    \item We introduce Adaptive Failure-Informed Learning (AFIL) for diffusion- and flow-based VLAs, using online-generated failure rollouts as a principled form of adaptive negative guidance for sequential action generation.
    \item We propose a Dual Action Generator (DAG)-VLA architecture that jointly learns from success and failure data while sharing a common VLM backbone, enabling efficient failure-aware policy learning with limited parameter overhead.
    \item We demonstrate that AFIL consistently improves success rates across diverse in-domain and out-of-domain manipulation tasks, spanning both short- and long-horizon settings, highlighting its effectiveness, efficiency, and generality.
\end{itemize}

\section{Related Work}
\label{sec:related_work}
\subsection{Robotic Manipulation via Imitation Learning}
Recent advances in robotic manipulation have widely adopted imitation learning, particularly behavior cloning, to address tasks ranging from grasping and pick-and-place to dexterous in-hand manipulation and assembly. Many prior works train deep visuomotor policies, typically based on convolutional neural networks, using simulation data or human demonstrations to enable efficient policy learning from visual observations. For example, \cite{DIL_ICRA18,Florence2019SelfSupervisedCI,pmlr-v15-ross11a,Avigal2022SpeedFoldingLE} employ explicit policy learning by directly regressing from the observed state to the action space. In contrast, another line of work formulates policy learning implicitly by modeling action distributions with Energy-Based Models (EBMs), allowing more expressive representations of multimodal behaviors \citep{florence2021implicit,LeCun2006ATO,pmlr-v119-grathwohl20a,Dai2019ExponentialFE}. We refer readers to \cite{RL_survey_AAAI2025} for a comprehensive survey of imitation- and reinforcement-learning-based methods for robotic policy learning.

\subsection{Vision-Language-Action (VLA) Models}
VLA models have emerged recently and demonstrated impressive generalization across diverse robotic tasks. Early large-scale systems such as RT-2 \citep{zitkovich2023rt}, PaLM-E~\citep{PaLM-E_ICML23} and Octo \citep{octo_2023} showed that policies conditioned on vision and language can follow semantic instructions, but they remain computationally expensive and closed-source. OpenVLA \citep{kim2024openvla} provides a powerful open-source alternative, leveraging large-scale cross-embodiment data and pretrained for improved generalist manipulation. 
Following~\cite{kim2024openvla}, numerous work have been proposed to improve visual action representation learning, \eg \cite{LaVA-Man-coRL25} introduces a self-supervised framework that learns robust visual-action representations by reconstructing masked goal images from language instructions. 
CoT-VLA \citep{zhao2025cot} and Embodied CoT \citep{zawalski2024robotic} incorporate Chain-of-Thought (CoT) reasoning and demonstrate that generating intermediate reasoning steps prior to action can improve task success and transparency. \cite{black2024pi_0,black2025pi_} introduces $\pi_0$/$\pi_{0.5}$, a general-purpose robot foundation model that combines internet-scale vision-language pre-training with a novel flow-matching architecture to enable versatile, dexterous manipulation across diverse robotic embodiments and complex, multi-stage tasks.
However, these approaches primarily learn from successful demonstrations. When the agent encounters an unexpected or erroneous state, VLA learned from success action space alone often fails to support recovery, as the reasoning process is not grounded in corrective experience. This limits its effectiveness in real-world settings where errors are inevitable.



\subsection{Learning from Failure in Robotic Policies} 
Failure signals are typically discarded during VLA training, despite containing valuable information about unsafe actions and recovery strategies. Existing efforts to leverage failures can be broadly categorized into three areas. \textbf{(i) Failure Detection and Precursors.} Several works focus on identifying unsafe or failure-prone states. SAFE \citep{gu2026safe} introduces a multitask failure detection framework based on internal VLA representations, while UNISafe \citep{seo2025uncertainty} proposes uncertainty-aware latent safety filters to prevent out-of-distribution failures. Other approaches identify precursors to failure using risk backpropagation or emergency-stop signals \citep{shangguan2025identifying}. While effective for prediction, these methods typically halt execution without providing goal-directed recovery and are often limited to narrow manipulation settings. \textbf{(ii) Interactive and Physical Correction.}
Human-in-the-loop approaches incorporate external feedback to correct failures during execution. \cite{shi2024yell} enables real-time language-based corrections, while \cite{zhang2024don} advocates physical intervention as a more natural interface. Although effective, these methods require continuous human supervision and do not scale to autonomous long-horizon deployment. \textbf{(iii) Autonomous Recovery and Optimization.}
Recent work explores learning from failure using optimization and reinforcement learning. Fail2Progress \citep{huang2025failprogress} applies Stein variational inference to learn from failures but relies on a fixed skill library. From Mystery to Mastery \citep{sagar2024mystery} combines RL-based exploration with VLM-based failure detection, though it focuses primarily on visual errors and offers limited insight into policy improvement. VLA with RL-based feedback, such as $\pi_{0.6}$ \citep{intelligence2025pi} or diagnostic systems like RoboFAC \citep{lu2025robofac}, demonstrate strong performance but require labor-intensive failure mode design and substantial in-domain data generation~\citep{lu2025robofac,lin2025failsafe,huang2025failprogress,duan2024aha}.

In contrast, our work targets data-efficient task completion for diffusion-/flow-based VLA models by providing two improvements: first we augment VLA training with failure-correction trajectories, enabling the policy to recover when failures occur. Second, improving sampling actions away from the failure manifold and sample in the failure-correction manifold when failure occurs.

\subsection{Classifier-Free Guidance for Guided Diffusion Sampling}
Classifier-Free Guidance (CFG) has demonstrated strong effectiveness in generative modeling \citep{ho2021classifierfree,Rombach2021HighResolutionIS,chung2025cfg,Fu_2025_ICCV,koulischer2025dynamic} by steering samples toward desired conditions through the contrast between conditional and unconditional predictions. Recent works have explored CFG-inspired strategies for guided diffusion sampling in control and policy generation settings \citep{reuss2023goal}. However, such approach relies on specialized numerical ODE solvers, making adaptation to existing pretrained VLA models nontrivial. More importantly, prior diffusion-based policy learning approaches do not explicitly exploit undesirable failure distributions, which provide informative counterexamples that can be leveraged through contrastive guidance objectives to further improve action generation.

Inspired by the idea of Negative Prompting for Text-to-Image (T2I) generation~\citep{gandikota2023erasing,schramowski2022safe,koulischer2025dynamic}, we propose a failure-informed VLA learning mechanism that adaptively biases action sampling toward regions of the embedding space associated with successful outcomes and away from regions associated with failures. Rather than relying on explicit planners or safety filters, our approach leverages the diffusion step-wise distance between success and failure distributions to implicitly guide action sampling away from failure modes and toward a confined success-oriented action space.
This allows the policy to remain flexible while avoiding failure-prone regions of the action manifold and recovering from unexpected states.

\section{Proposed Methodology}
\label{sec:method}
\newcommand{\E}{\mathbb{E}}
\newcommand{\R}{\mathbb{R}}
\newcommand{\KL}{\mathrm{KL}}
\newcommand{\vx}{\bm{x}}
\newcommand{\va}{\bm{a}}
\newcommand{\vo}{\bm{o}}
\newcommand{\vell}{\bm{\ell}}

\begin{figure*}[!t]
    \centering
    \includegraphics[width=\textwidth]{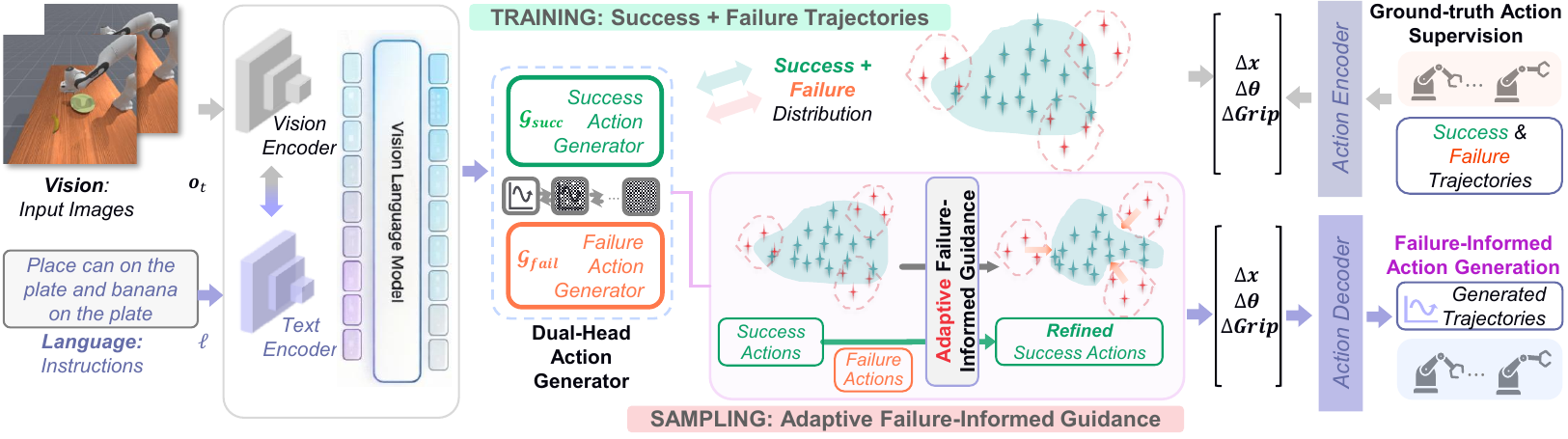}
    \caption{Dual Action Generator (DAG)-VLA with Adaptive Failure-Informed Learning (AFIL).}
    \label{fig:pipeline} 
\end{figure*}

In this section we describe our approach for Adaptive Failure-Informed Learning (AFIL) based on diffusion- and flow-based VLA models. Our method leverages two action generators: one trained on successful trajectories and another trained on failure trajectories. 
At inference/sampling time, the failure generator provides adaptive negatively prompted guidance that steers action sampling away from failure-prone regions of the action space. We first describe the dual action generator (DAG) architecture and training strategy, followed by adaptive failure-informed sampling mechanism. Please refer to Figure~\ref{fig:pipeline} for detailed illustration of the proposed AFIL pipeline.

\subsection{Preliminaries}
\label{sec:problem}
\textbf{Vision-Language-Action (VLA) Models.} We aim to learn vision-language-action (VLA) policies, $\pi_{\theta}(\va_t \mid \vo_t,\ell)$, that generate continuous robot actions conditioned on visual observations and language instructions. At time step $t \in {1, \dots, T}$, where $T$ denotes the length of a trajectory, the policy receives an observation $\vo_t = (I_t, \vx_t)$, where $I_t$ denotes one or more RGB (or RGB-D) images and $\vx_t$ denotes proprioceptive states, together with a language instruction $\ell$. The policy outputs an action chunk $\va_{t:t+H} \in \mathbb{R}^{d \times H}$, where $\va_t^{d}$ is the $d$-DoF robot joint/gripper control vector (including joint states, joint angles and gripper states) at time $t$, $H$ is the action horizon. VLA models \citep{kim2024openvla,black2025pi_,black2024pi_0,LaVA-Man-coRL25} typically consist of a large-scale vision-language model (VLM) backbone $\Phi_{\text{VLM}}(I_t,\ell)$ followed by a specialized action generation module $\mathcal{G}$. In this work, we focus on diffusion- and flow-based VLAs \citep{black2025pi_,black2024pi_0,LaVA-Man-coRL25} due to their strong ability to model multimodal action distributions. 

\textbf{Classifier-Free Guidance}
Classifier-free guidance (CFG) \citep{chung2025cfg} jointly train a diffusion model to handle both conditional and unconditional generation, moving the score estimate toward the conditional prediction to trade off sample diversity for higher visual fidelity and better command following. Specifically, \cite{chung2025cfg} applies Bayes rule to rewrite the sharpened posterior distribution $p_\eta(\mathbf{x}|\mathbf{c})$ at diffusion step $\eta$, given condition $\mathbf{c}$:
\begin{align}
\nabla_{\mathbf{x}} \log p_\eta(\mathbf{x}|\mathbf{c})
= \nabla_{\mathbf{x}} \log p_\eta(\mathbf{x}) + \lambda (\nabla_{\mathbf{x}} \log p_\eta(\mathbf{x}|\mathbf{c}) - \nabla_{\mathbf{x}} \log p_\eta(\mathbf{x}))
\label{eq:cfg_p}
\end{align}

Parameterizing with the score function~\citep{DDPM_NeurIPS20}, the classifier free-guided score $\mathbf{s}_{\text{CFG}}^*$ is a linear combination of unconditional score $s_{uc}$ and conditional score $s_{c}$:
\begin{align}
\mathbf{s}_{\text{CFG}}^*
= s_{\mathbf{uc}} + \lambda (s_{\mathbf{c}} - s_{\mathbf{uc}})
\label{eq:cfg_score}
\end{align}

For Negative Prompting (NP)~\citep{gandikota2023erasing,schramowski2022safe,koulischer2025dynamic}, the posterior $\log p_t(\mathbf{x}| \mathcolor{darkgreen}{\mathbf{c}^+})$ ($\mathcolor{darkgreen}{\mathbf{c}^+}$, and $\mathcolor{red}{\mathbf{c}^-}$ denote positive and negative prompt respectively):
\begin{align}
\nabla_{\mathbf{x}} \log p_t(\mathbf{x}| \mathcolor{darkgreen}{\mathbf{c}^+} )
= \nabla_{\mathbf{x}} \log p_t(\mathbf{x}) - \lambda (\nabla_{\mathbf{x}} \log p_t(\mathbf{x}|\mathcolor{red}{\mathbf{c}^-}) - \nabla_{\mathbf{x}} \log p_t(\mathbf{x}))
\label{eq:ng_p}
\end{align}
The NP guided score $\mathbf{s}^*_{\text{NP}}$ linearly combines negative prompt score $s_{\mathcolor{red}{\mathbf{c}^-}}$ and unconditional score $s_{\mathbf{uc}}$:
\begin{align}
\mathbf{s}_{\text{NP}}^*
= s_{\mathbf{uc}} - \lambda (s_{\mathcolor{red}{\mathbf{c}^-}} - s_{\mathbf{uc}})
\label{eq:ng_score}
\end{align}
The static guidance scale can be further set to $\lambda(\mathbf{x}, \eta) \propto \frac{p_\eta(\mathbf{\mathcolor{red}{\mathbf{c}^-}|\mathbf{x}})}{1-p_\eta(\mathbf{\mathcolor{red}{\mathbf{c}^-}|\mathbf{x}})}$ values for state-dependent guidance~\citep{koulischer2025dynamic}.

\subsection{Dual Action Generator (DAG) Architecture}
\label{sec:DAG}
In conventional success-only VLA training, the policy models an action distribution $p_{\text{succ}}(\va \mid \vo,\ell)$ using large-scale successful robot demonstrations and trajectories \citep{o2024open, black2025pi_}. However, failure trajectories also contain valuable information about action patterns that lead to unsuccessful outcomes. 
We introduce an additional lightweight action generator that operates in parallel with $\mathcal{G}$, extending conventional VLA architectures to explicitly model both success and failure action distributions. 
Let $p_{\text{fail}}(\va \mid \vo,\ell)$ denotes the distribution of actions associated with failure outcomes. Our objective is to (i) learn both success and failure action manifolds, and (ii) leverage them during inference to guide action sampling away from failure.

Specifically, our proposed dual action generator (DAG) VLA composes of,
(i) a \emph{Success} action generator $\mathcal{G}_{\text{succ}}$ trained only on successful trajectories, and (ii) a \emph{Failure} action generator $\mathcal{G}_{\text{fail}}$ trained only on failure trajectories. Both generators share the same vision-language backbone $\Phi_{\text{VLM}}$. Given vision-language features $\mathbf{h}=\Phi_{\text{VLM}}(\vo,\ell)$, the generators induce two generative action models, $p_{\theta_s}(\va\mid \vo,\ell)$ and $p_{\theta_f}(\va\mid \vo,\ell)$, parameterized by their action-head parameters $\theta_s$ and $\theta_f$, respectively.

\textbf{Failure Data Generation.}
Existing approaches for leveraging mixed success–failure datasets often rely on handcrafted failure taxonomies~\citep{duan2024aha,lin2025failsafe,huang2025failprogress}. For example, prior work categorizes errors such as unstable grasps or kinematic slips to guide synthetic data generation~\citep{duan2024aha}, but such heuristics are inherently limited in capturing the stochastic and diverse nature of real-world failures. In contrast, we incorporate online-inferred failure trajectories obtained directly from rollouts of a learned or pretrained VLA under task-conditioned execution. This design provides two advantages: it captures a richer, more realistic failure distribution, including emergent and compounding error modes, and it eliminates the need for manual annotation or handcrafted perturbation design by leveraging naturally occurring rollout failures. Consequently, the resulting failure data better approximates the on-policy state-action distribution, improving robustness and generalization. These failure trajectories can further be corrected via motion-planner-based replanning, enabling effective augmentation of success-only training data and improving overall VLA performance (see Section~\ref{sec:exp}).

\textbf{Training Objectives.} Let training dataset $\mathcal{D} = \{\mathcal{D}_s, \mathcal{D}_f\}$, where $\mathcal{D}_s$ and $\mathcal{D}_f$ contain success and failure trajectories respectively. 
We train each action generator on its respective dataset using the same generative objective but with disjoint supervision. For diffusion-based policies~\citep{DDPM_NeurIPS20,black2024pi_0}, each action generator predicts noise $\epsilon_{\theta}(\va^\eta,\vo,\ell,\eta)$ at diffusion step $\eta$ and minimizes,

\begin{align}
\mathcal{L}_{\text{diff}}(\theta; \mathcal{D})
=
\mathbb{E}_{(\vo,\ell,\va)\sim\mathcal{D}}
\;
\mathbb{E}_{\eta,\epsilon}
\left[
\left\|
\epsilon - \epsilon_{\theta}(\va^\eta,\vo,\ell,\eta)
\right\|_2^2
\right].
\end{align}

Alternatively, for flow-based policies~\citep{black2025pi_} the action head predicts a velocity field
$v_{\theta}(\va^\eta,\vo,\ell,\eta)$ that transports samples from noise to the data distribution. The model is trained to match the ground-truth velocity between noisy actions and target actions using the standard flow-matching objective. Our method does not modify underlying generative VLA training objective. 

\subsection{Adaptive Failure-Informed Sampling}
\label{sec:FIL}

At inference, naive sampling from $p_{\theta_s}$ can propose actions that drift into failure-prone regions. Inspired by Negative Prompting for Text-to-Image (T2I) generation~\citep{gandikota2023erasing,schramowski2022safe,koulischer2025dynamic}, we propose to perform \emph{adaptive failure-informed sampling}, given the learned success and failure distributions from the DAG (Sec.~\ref{sec:DAG}) to fully explore the repulsive nature of the undesired distributions, which is ignored in conventional VLA learning.

Let $\epsilon_{\textcolor{darkgreen}{\text{succ}}}(\va^\eta,\vo,\ell,\eta)$ denote the predicted noise from the success generator $\mathcal{G}_{\text{succ}}$ and $\epsilon_{\textcolor{red}{\text{fail}}}(\va^\eta,\vo,\ell,\eta)$ denotes the predicted noise by the failure action generator at diffusion time step $\eta$. Here we use diffusion models for illustration purpose. For flow-matching models, $\epsilon_{\textcolor{darkgreen}{\text{succ}}}(\va^\eta,\vo,\ell,\eta)$ and $\epsilon_{\textcolor{red}{\text{fail}}}(\va^\eta,\vo,\ell,\eta)$ can be exchanged with velocity field $v_{\textcolor{darkgreen}{\text{succ}}}(\va^\eta,\vo,\ell,\eta)$ and $v_{\textcolor{red}{\text{fail}}}(\va^\eta,\vo,\ell,\eta)$ respectively. In the following, we omit the explicit conditioning on $(\vo,\ell)$ and the timestep argument $\eta$, and write $\epsilon_{\textcolor{darkgreen}{\text{succ}}}(\va^\eta)$ and $\epsilon_{\textcolor{red}{\text{fail}}}(\va^\eta)$ for notational simplicity.

Recall Equation~\ref{eq:ng_score}, Negative Prompting (NP) guidance can be applied to steer the diffusion sampling more accurately towards desired positive distributions by linearly combining negative prompt score and unconditional score. For the proposed DAG-VLA, we have failure-informed (FI) score with \emph{adaptive guidance} to steer the diffusion sampling away from failure:

\begin{gather}
\mathbf{\epsilon}^*_{\text{FI}}
=
\epsilon_{\textcolor{darkgreen}{\text{succ}}}(\va^\eta, \varnothing)
-
\lambda_\eta(\va^\eta) \,
\left(
\epsilon_{\textcolor{red}{\text{fail}}}(\va^\eta)
-
\epsilon_{\textcolor{darkgreen}{\text{succ}}}(\va^\eta, \varnothing)
\right)
\\
\Leftrightarrow
\epsilon_{\textcolor{darkgreen}{\text{succ}}}(\va^\eta, \varnothing)
-
\hat{\lambda}_\eta(\va^\eta)
\epsilon_{\textcolor{red}{\text{fail}}}(\va^\eta)
\label{eq:dfil_score}
\end{gather}

Diffusion guidance in Equation~\ref{eq:dfil_score} approximates score combination: $\nabla_{\va} \log p_{\textcolor{darkgreen}{\text{succ}}}(\va^\eta) - \hat{\lambda}_\eta(\va^\eta) \, \nabla_{\va} \log p_{\textcolor{red}{\text{fail}}}(\va^\eta)$, we choose $\hat{\lambda}_\eta(\va^\eta)$ to be proportional to success-failure distribution distance $D(p_{\textcolor{darkgreen}{\text{succ}}}, p_{\textcolor{red}{\text{fail}}})$, where $D(\cdot \,, \cdot)$ is the divergence metric between two distributions. 

Intuitively, this quantity captures how distinguishable successful and failed behaviors are in the local state: when the two scores are highly aligned, indicating that success and failure modes are locally ambiguous, strong negative guidance may suppress valid actions. Conversely, when the scores are dissimilar or oppositely aligned, the failure model provides complementary information about undesirable directions, making stronger negative guidance beneficial for steering the trajectory away from failure modes. 
This adaptive mechanism is particularly important in robotic manipulation, where successful and failed trajectories may share overlap before diverging. \Eg, at the beginning of an episode, both successful and failed rollouts may move from the home pose toward the target object using similar reaching motions. In such states, applying strong failure repulsion would be undesirable, the guidance scale should remain small. In contrast, once a clear failure mode emerges, such as dropping, unstable grasping, or drifting away from the target, the success and failure predictions become more distinguishable. At these stages, increasing the guidance scale allows the policy to actively suppress failure-prone actions and bias sampling toward corrective behaviors. This yields a simple, fully local, and computation-efficient mechanism that adaptively balances attraction toward successful behaviors and repulsion from failure modes only when such correction is necessary.

Specifically we adopt a cosine similarity-based strategy to adaptively modulate the strength of negative guidance during diffusion sampling, \ie, 
\begin{align}
\hat{\lambda}_\eta(\va^\eta) = \alpha (\, 1-\cos(\epsilon_{\textcolor{darkgreen}{\text{succ}}}, \epsilon_{\textcolor{red}{\text{fail}}}) \, ),
\label{eq:dynamic_lambda}
\end{align}
where $\alpha$ is the guidance scale. While several metrics can quantify the divergence between success and failure modes, we found the cosine distance to be uniquely suited for the adaptive weighting of VLA diffusion scores, considering several aspects: a) Distribution-based metrics such as Maximum Mean Discrepancy (MMD) or Jensen-Shannon Divergence (JSD) provide theoretically robust measures of divergence but require processing batches of trajectories at each denoising step. For real-time VLA applications, the overhead of calculating kernel matrices or density estimations at every iteration $\eta$ is computationally prohibitive. b) In robotics and motion synthesis, the "semantic" intent of an action is encoded more heavily in the angular orientation of the joint-space or end-effector vector than in its absolute coordinate-wise difference. Metrics like Manhattan ($L_1$) or Chebyshev distance treat each dimension independently and are sensitive to the coordinate system's scaling. Cosine distance, by contrast, captures the global "intent" of the motion, muting the guidance when the success and failure trajectories are semantically aligned and amplifying it only when they represent distinct behavioral choices. Please see supplementary material for mathematical justifications.






\section{Evaluation}
\label{sec:eval}
\label{sec:exp}
To evaluate the proposed AFIL framework, we conduct experiments on both in-domain and out-of-domain manipulation tasks across multiple simulation environments, including ManiSkill~\citep{tao2025maniskill, gu2023maniskill2} and LIBERO~\citep{liu2023libero}. We additionally provide real-robot validation on a tabletop manipulation task with a Trossen ViperX 300S 6-DoF robotic arm. 
All experiments are built upon the $\pi_{0.5}$ architecture~\citep{black2025pi_}; however, AFIL is model-agnostic and can be readily integrated with diffusion- or flow-matching-based VLA policies. See supplementary material for additional implementation, experiment details and parameter/runtime analysis.

\subsection{Environment Setups}
\label{sec:setup}
We evaluate DAG-VLA under two experimental settings:

\textbf{i) Maniskill}~\citep{tao2025maniskill}: 
We perform both in-domain and out-of-domain evaluations in the ManiSkill environment using a Franka Emika Panda 7-DoF robot arm. For in-domain evaluation, we consider three short-horizon tasks - \textit{stack red cube on green cube}, \textit{lift the peg upright}, and \textit{pull cube to the red and white target} — as well as two long-horizon tasks: \textit{place bowl on the plate and spoon on the bowl}, and \textit{place can on the plate and banana on the plate}. 
We train our DAG-VLA on 1500 success, 1000 failure-corrected, and 1500 failure trajectories. For in domain evaluation, we evaluate DAG-VLA on 50 initialization configurations (varying initial states, object/robot poses) with 3 runs for each configuration, \ie 50$\times$3=150 rollouts, and report the absolute success rate for each task.

\begin{figure*}[h!]
    \centering
    \includegraphics[width=\textwidth]{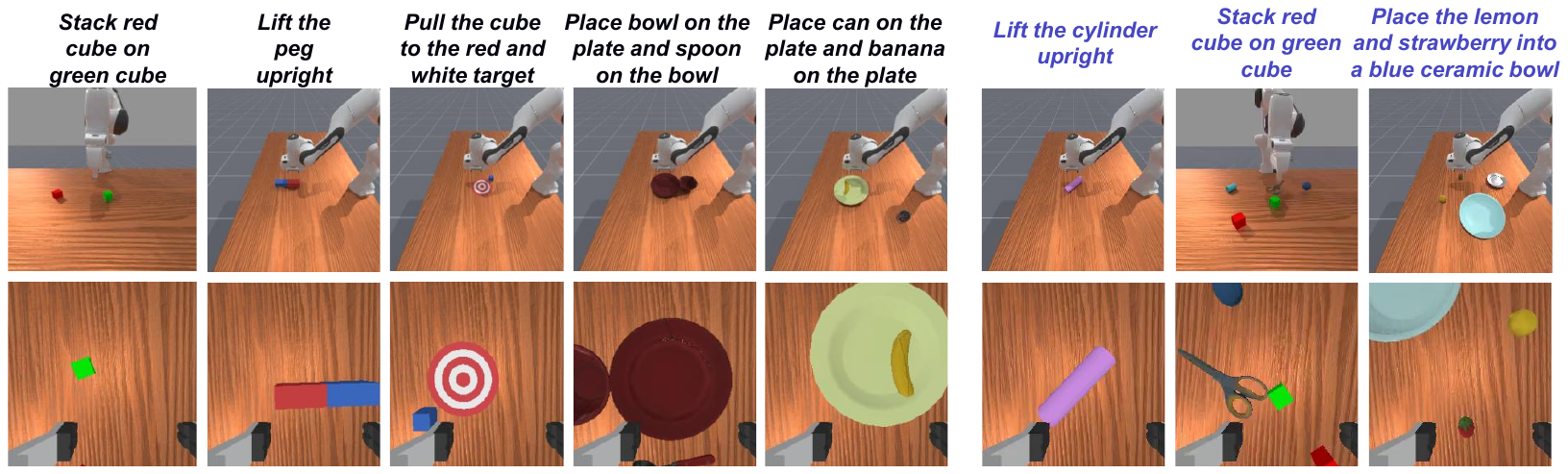}
    \caption{Visualization of in-domain (left) and out-of-domain (right) manipulation task setup. Row 1: exterior camera image, row 2: wrist camera image. See supplementary for more visualizations.}
    \label{fig:setup_vis} 
\end{figure*}

For out-of-domain evaluation, we assess generalization to unseen objects with different geometries and textures in cluttered tabletop scenes. Specifically, we evaluate two short-horizon tasks: \textbf{Lift Cylinder} (\textit{lift the cylinder upright}, with unseen geometry and color) and \textbf{Stack Cubes} (\textit{stack red cube on green cube}, in a cluttered scene with distractors), along with one long-horizon task, \textbf{Fruit Sorting} (\textit{place the lemon and strawberry into a blue ceramic bowl}, with unseen objects and color). For out-of-domain tasks, we evaluate on 25 (configs)$\times$3 (runs) = 75 rollouts for each task. Please refer to Figure~\ref{fig:setup_vis} for visualizations of the task setups in Maniskill.

\textbf{ii) LIBERO}~\citep{liu2023libero}: To enable fair comparison with prior work, we follow the standard LIBERO benchmark evaluation protocol~\citep{kim2024openvla, black2025pi_}. The benchmark comprises four task suites, each containing 10 tasks with 50 human-teleoperated demonstrations. These suites capture complementary generalization challenges: \textit{LIBERO-Spatial} (varying layouts), \textit{LIBERO-Object} (varying object identities), \textit{LIBERO-Goal} (varying task objectives), and \textit{LIBERO-Long (LIBERO-10)} (long-horizon tasks with diverse objects, layouts, and goals).
We collect failure trajectories from pretrained $\pi_{0.5}$ (finetuned on LIBERO) rollout executions and use them to train the proposed DAG-VLA model. We report success rate as the average over 3 random seeds x 500 rollouts each (10 tasks $\times$ 50 rollouts per task) following~\cite{kim2024openvla}.

\subsection{In-Domain Evaluation}
\subsubsection{Manipulation Tasks in Maniskill}
In Table~\ref{tab:maniskill_eval}, we present success rates of proposed DAG-VLA under various ablated settings. Success-only follows standard VLA training, where $\pi_{0.5}$ is trained solely on successful demonstration trajectories. Success + Failure Correction augments this data by first executing the success-only policy to collect failure cases, which are then corrected via motion-planner-based replanning and used for fine-tuning. Building on this, Static-FIL trains the proposed DAG-VLA using both success data and (corrected and raw) failure trajectories, and applies failure-informed sampling at inference with a fixed guidance strength $\lambda$ (empirically set to 0.05 which works best for this setting; see supplementary material for ablations). Finally, Adaptive-FIL uses the same DAG-VLA training data but replaces the static weighting with adaptive failure-informed sampling (Eq.~\ref{eq:dfil_score} and Eq.~\ref{eq:dynamic_lambda}), where the guidance scale is adaptively adjusted during rollout based on failure signals. Guidance strength $\alpha$ (\cf Eq.~\ref{eq:dynamic_lambda}) is set to 1.0, please see next section for ablation studies.

Across both short- and long-horizon manipulation tasks, the proposed DAG-VLA with Adaptive Failure-Informed Learning (Adaptive-FIL) consistently achieves the strongest performance. Compared to training with success-only demonstrations, incorporating failure correction already yields substantial gains, particularly on less challenging tasks such as \textit{Pull cube to the red and white target}. Introducing static failure-informed sampling further improves results, demonstrating the benefit of leveraging failure signals during inference. However, Adaptive-FIL provides the most consistent and significant improvement across all tasks, outperforming both static sampling and all baselines. Notably, it achieves the largest gains on long-horizon tasks, where compounding errors are more prevalent (e.g., +7.4\% and +12.0\% over static-FIL on the two long-horizon settings). These results indicate that dynamically adapting failure guidance during rollout is crucial for robust long-horizon, multi-step manipulation and generalization in complex environments.

\begin{table*}[h!]
\centering
\small
\renewcommand{\arraystretch}{1.2}
\caption{Evaluation and ablation study of proposed DAG-VLA on various manipulation tasks in Maniskill environments. For each task, success rates (\%) are evaluated over 50 configs $\times$ 3 runs.}
\label{tab:maniskill_eval}
\resizebox{\textwidth}{!}{%
\begin{tabular}{c|l|C{1.5cm}C{1.5cm}C{1.8cm}C{2cm}}
\toprule
\multicolumn{2}{c|}{\textbf{TASK}} &
\textbf{Success Only} & \textbf{Success + Failure Correction} & \textbf{DAG-VLA + Static-FIL ($\lambda=0.05$)} &
\textbf{DAG-VLA + Adaptive-FIL ($\alpha=1.0$)} \\
\midrule
\multirow{3}{*}{\parbox{1.2cm}{\centering\textbf{Short Horizon}}} & \textit{Stack red cube on green cube} & 71.3 & 76.7 & 80.0 & \textbf{84.7}\\
& \textit{Lift the peg upright} & 70.7 & 80.0 & 86.7 & \textbf{90.7} \\
& \textit{Pull cube to the red and white target} & 60.0 & 92.0 & 94.7 & \textbf{98.0} \\[1pt]
\midrule
\multirow{2}{*}{\parbox{1.2cm}{\centering\textbf{Long Horizon}}} & \textit{Place bowl on the plate and spoon on the bowl} &
38.6 & 45.3 & 49.3 & \textbf{56.7} \\
& \textit{Place can on the plate and banana on the plate} &
39.3 & 56.0 & 59.3 & \textbf{71.3} \\
\bottomrule
\end{tabular}
}
\end{table*}

\textbf{Ablation Study on Guidance Scale}
We conduct an ablation study to examine how the scale of the adaptive guidance, $\alpha$ (\cf Eq.~\ref{eq:dynamic_lambda}), affects performance on two manipulation tasks: \textit{stack red cube on green cube}, a short-horizon task, and \textit{place can on the plate and banana on the plate}, in Table~\ref{tab:abl_alpha}.

\begin{table*}[h!]
\centering
\small
\caption{Ablation study on guidance scale $\alpha$ for AFIL. Success rates reported in \%.}
\label{tab:abl_alpha}
\renewcommand{\arraystretch}{1.2}
\resizebox{.8\textwidth}{!}{%
\begin{tabular}{l|cccc}
\toprule
\textbf{Task} &
$\alpha$ = 0.5 &
$\alpha$ = 1.0 &
$\alpha$ = 2.0 &
$\alpha$ = 5.0 \\
\toprule
\textit{Stack red cube on green cube} & 84.0 & \textbf{84.7} & 79.3 & 76.0 \\
\textit{Place can on the plate and banana on the plate} & 68.7 & \textbf{71.3} & 69.3 & 66.0 \\
\bottomrule
\end{tabular}%
}
\end{table*}

\textbf{Ablation Study on Failure Correction Data} Failure correction and failure-informed guidance serve distinct roles. Corrected trajectories augment the positive distribution with recovery behaviors, whereas raw failure trajectories train the failure generator and provide negative guidance. AFIL does not mathematically require corrected trajectories: the success generator can be trained on the original successful demonstrations, while the failure generator is trained on raw rollout failures. 

We therefore perform an ablation comparing DAG-VLA trained with and without corrected trajectories under the AFIL evaluation setting on one short-horizon task and one long-horizon task in Table~\ref{tab:failure_correction_afil}. The results show consistent improvement of our introduced AFIL (adaptive negative guidance) with and without failure correction data.


\begin{table*}[h!]
\centering
\caption{Success rates (\%) with and without AFIL under different failure-correction settings.}
\label{tab:failure_correction_afil}
\small
\setlength{\tabcolsep}{10pt}
\resizebox{.9\textwidth}{!}{%
\begin{tabular}{l|cc|cc}
\toprule
\multicolumn{1}{c|}{\multirow[c]{2}{*}{\textbf{TASK}}} & \multicolumn{2}{c|}{\textbf{No Failure Correction}}
& \multicolumn{2}{c}{\textbf{With Failure Correction}} \\
\cmidrule(lr){2-3} \cmidrule(lr){4-5}

& w/o AFIL & \textbf{w/ AFIL}
& w/o AFIL & \textbf{w/ AFIL} \\
\midrule
\textit{Lift the Peg Upright}
& 71.3 & \textbf{81.3}
& 76.7 & \textbf{90.7} \\

\textit{Place Bowl on Plate and Spoon on Bowl}
& 38.6 & \textbf{50.7}
& 45.3 & \textbf{56.7} \\
\bottomrule
\end{tabular}
}
\end{table*}

\subsubsection{Benchmark Evaluation on LIBERO}

In Table~\ref{tab:libero_results}, we compared proposed DAG-VLA ($\pi_{0.5}$ backbone) with existing VLA backbones on LIBERO benchmark. Across the LIBERO benchmark suites, DAG-VLA with AFIL consistently achieves the strongest performance among all methods. Compared to prior VLA models - including Diffusion Policy, Octo, OpenVLA, CoT-VLA, and a strong $\pi_{0.5}$ baseline - our method improves performance across all four task suites, with particularly notable gains on the most challenging LIBERO-10 \emph{long-horizon} setting. While $\pi_{0.5}$ already provides a strong baseline, incorporating AFIL further improves average performance from 96.9\% to 98.4\%, indicating that even strong pretrained policies benefit from adaptive failure-informed guidance. These results highlight the effectiveness of AFIL in improving robustness and long-horizon generalization by dynamically adjusting failure guidance during inference, rather than relying on fixed or purely success-driven rollouts.

\begin{table*}[h!]
\centering
\small
\caption{Performance comparison on LIBERO~\citep{liu2023libero} benchmark. Success rates in \%.}
\label{tab:libero_results}
\resizebox{\textwidth}{!}{%
\begin{tabular}{l|C{1.5cm}C{1.5cm}C{1.5cm}C{1.5cm}|M{1.5cm}}
\toprule
\textbf{Method} &
\textbf{LIBERO Spatial} &
\textbf{LIBERO Object} &
\textbf{LIBERO Goal} &
\textbf{LIBERO 10} &
\textbf{Average} \\
\toprule
Diffusion Policy from scratch~\citep{chi2025visuomotor} &78.3 $\pm$ 1.1 & 92.5 $\pm$ 0.7 & 68.3 $\pm$ 1.2 & 50.5 $\pm$ 1.3 & 72.4 $\pm$ 0.7 \\
Octo fine-tuned~\citep{octo_2023} &
78.9 $\pm$ 1.0 & 85.7 $\pm$ 0.9 & 84.6 $\pm$ 0.9 & 51.1 $\pm$ 1.3 & 75.1 $\pm$ 0.6 \\
OpenVLA fine-tuned~\citep{kim2024openvla} &
84.7 $\pm$ 0.9 & 88.4 $\pm$ 0.8 & 79.2 $\pm$ 1.0 & 53.7 $\pm$ 1.3 & 76.5 $\pm$ 0.6 \\
CoT-VLA-7B~\citep{zhao2025cot} & 87.5 $\pm$ 1.4 & 91.6 $\pm$ 0.5 & 87.6 $\pm$ 0.6 & 69.0 $\pm$ 0.8 & 81.1 $\pm$ 0.6 \\
$\pi_{0.5}$~\citep{black2025pi_} &
98.8 $\pm$ na & 98.2 $\pm$ na & 98.0 $\pm$ na & 92.4 $\pm$ na & 96.9 $\pm$ na \\
\midrule
$\text{DAG-VLA}_{\pi_{0.5}}$ + \textbf{AFIL} &
\textbf{99.8 $\pm$ 0.4} & \textbf{99.7 $\pm$ 0.8} & \textbf{98.1 $\pm$ 0.4} & \textbf{95.8 $\pm$ 1.2} & \textbf{98.4 $\pm$ 0.7} \\
\bottomrule
\end{tabular}%
}
\end{table*}

\subsection{Out-of-Domain (OOD) Evaluation}

For OOD evaluation, we test generalization to unseen object instances, novel geometries/textures, and cluttered tabletop environments (Table~\ref{tab:ood_results}). The success-only $\pi_{0.5}$ policy degrades substantially under these distribution shifts, while incorporating corrected failure data improves performance across all tasks. Our $\text{DAG-VLA}_{\pi_{0.5}}$ with AFIL achieves the best results in all settings, showing consistent gains on both short- and long-horizon tasks. These results demonstrate that adaptive failure-informed guidance improves robustness to visual, geometric, and task-level distribution shifts.

\begin{table*}[h!]
\centering
\caption{Performance comparison on out-of-domain manipulation tasks. Success rates (\%) reported over 25 configs $\times$ 3 runs for each task.}
\label{tab:ood_results}
\renewcommand{\arraystretch}{1.2}
\resizebox{.9\textwidth}{!}{%
\begin{tabular}{l|C{3cm}C{3cm}|C{4cm}}
\toprule
\multirow{2}{*}{\large{\textbf{Method}}} & \multicolumn{2}{c|}{\textbf{Short Horizon}} & \textbf{Long Horizon} \\[1pt]
\cline{2-4}\noalign{\vskip 2pt}
& \makecell{\textbf{Lift Cylinder} \\ (unseen object/color)} &
\makecell{\textbf{Stack Cubes} \\ (cluttered scene)} &
\makecell{\textbf{Fruit Sorting} \\ (unseen objects + clutter)} \\
\toprule
$\pi_{0.5}$ (success only) & 14.7 & 16.0 & 10.7 \\
\midrule
$\pi_{0.5}$ (success + failure correction) & 53.3 & 41.3 & 54.7 \\
$\text{DAG-VLA}_{\pi_{0.5}}$ + \textbf{AFIL} & \textbf{62.7} & \textbf{48.0} & \textbf{64.0} \\
\bottomrule
\end{tabular}%
}
\end{table*}




\subsection{Real-Robot Validation}
To examine whether the benefit of failure-informed guidance persists beyond simulation, we conduct a controlled proof-of-concept evaluation on a Trossen ViperX 300S 6-DoF robotic arm using the tabletop manipulation task, \textit{Pick up the yellow cube}. We collected 50 successful and failure-corrected trajectories and 50 raw failure trajectories to fine-tune DAG-VLA from the base $\pi_{0.5}$ checkpoint. We then evaluated the same trained DAG-VLA model with and without adaptive negative guidance (AFIL) over 50 real-robot trials, covering five initial cube configurations with 10 trials per configuration. The success rates are reported in Table~\ref{tab:realrobot}. 
AFIL improves the average real-robot success rate from 0.50 to 0.92, and improves performance across all five initial configurations. These results provide real-world evidence that the learned failure-conditioned guidance transfers beyond simulation and can improve robustness under physical perception, contact, and actuation uncertainties. Please see supplementary material for workspace setup, qualitative results, and rollout visualizations.

\begin{table*}[h!]
\centering
\caption{Performance comparison on real-world tabletop manipulation task. Success rates (0.00-1.00) are evaluated over 5 configurations with 10 trials per configuration.
}
\renewcommand{\arraystretch}{1.2}
\resizebox{.85\textwidth}{!}{%
\begin{tabular}{l|cccccc}
\toprule
Method & Config. 0 & Config. 1 & Config. 2 & Config. 3 & Config. 4 & Average \\
\midrule
$\text{DAG-VLA}_{\pi_{0.5}}$,~{No FIL} & 0.20 & 0.00 & 0.70 & 0.90 & 0.70 & 0.50 \\
\textbf{$\text{DAG-VLA}_{\pi_{0.5}}$ + AFIL} & \textbf{1.00} & \textbf{0.80} & \textbf{0.90} & \textbf{1.00} & \textbf{0.90} & \textbf{0.92} \\
\bottomrule
\end{tabular}
}
\label{tab:realrobot}
\end{table*}

\section{Conclusion}
\label{sec:conclusion}
We introduced Adaptive Failure-Informed Learning (AFIL), a framework for improving diffusion- and flow-based VLA policies by using failure trajectories as adaptive negative guidance. AFIL augments conventional success-only VLA learning with a Dual Action Generator architecture that jointly models successful and failed behaviors while sharing a common vision-language backbone. During inference, the failure generator adaptively guides action sampling away from failure-prone regions, improving robustness without requiring handcrafted failure modes, task-specific recovery heuristics, or separate large-scale VLA models. Across ManiSkill and LIBERO evaluations, AFIL consistently improves success rates over strong VLA baselines, with particularly notable gains on long-horizon and out-of-domain manipulation tasks. These results show that explicitly modeling failure distributions can provide an efficient and scalable mechanism for robust action generation. We believe failure-informed guidance offers a promising direction for developing VLA policies that better generalize to realistic deployment settings where errors, distribution shifts, and recovery demands are unavoidable.

\bibliographystyle{plainnat}
\bibliography{references}

\newpage
\renewcommand{\thetable}{S\arabic{table}}
\setcounter{table}{0}

\clearpage
\setcounter{section}{0}


\begin{center}
    {\Large \bfseries Supplementary Material}
\end{center}

\section{Implementation Details}
\label{sec:implementation}
We augment the $\pi_{0.5}$ VLA architecture with Dual Action Generator (DAG), training the proposed $\text{DAG-VLA}_{\pi_{0.5}}$ using both success and failure trajectories (\cf Section~\ref{sec:DAG}), and applying failure-informed sampling during inference (\cf Section~\ref{sec:FIL}) for action rollout. 
During sampling, we use positively prompted noise $\mathbf{v}_{\textcolor{darkgreen}{\text{succ}}}(\va^\eta)$ to replace unguided model $\mathbf{v}_{\textcolor{darkgreen}{\text{succ}}}(\va^\eta, \varnothing)$ (Eq.~\ref{eq:dfil_score}) following~\cite{schramowski2022safe,koulischer2025dynamic}.

Data preprocessing, including conversion to the LeRobot format~\citep{cadenelerobot}, and normalization follow~\cite{black2025pi_}. We set the action dimension to 32 and the action horizon to 30 steps. Training is initialized from the public $\pi_{0.5}$ base checkpoint and runs for 300k steps with a batch size of 4. All experiments are conducted on 4 NVIDIA A100-SXM4-80GB GPUs, with a total training time of approximately 8 hours.

\section{Qualitative Results}
Please refer to the supplementary videos for qualitative results. For in-domain and out-of-domain evaluations in the ManiSkill environment (corresponding to Section 4.2.1 and Section 4.2.3 in the main paper), we visually compare the proposed Dual Action Generator (DAG)-VLA under three settings: training with success-only data, training with success plus failure-correction data, and with Adaptive Failure-Informed Learning (AFIL). Quantitative comparisons are reported in Table~1 of the main paper. The videos show that VLAs trained only on successful demonstrations, following the conventional setting, are prone to failure and lack effective recovery capability. Incorporating failure-correction data mitigates this issue, while DAG-VLA with AFIL enables more accurate failure avoidance and correction, leading to higher task success rates.

For evaluation in the LIBERO environment, we follow the protocols of~\cite{kim2024openvla,zhao2025cot,black2025pi_} and present qualitative comparisons between the official $\pi_{0.5}$ checkpoint fine-tuned on LIBERO\footnote{Official checkpoint released at: https://github.com/Physical-Intelligence/openpi/blob/main/examples/libero/README.md} and our proposed DAG-VLA with AFIL built upon the $\pi_{0.5}$ backbone. We include representative tasks from the LIBERO Spatial, Object, Goal, and Long test suites.

\begin{figure}[h!]
    \centering

    \begin{subfigure}[t]{\textwidth}
        \centering
        \includegraphics[width=\linewidth]{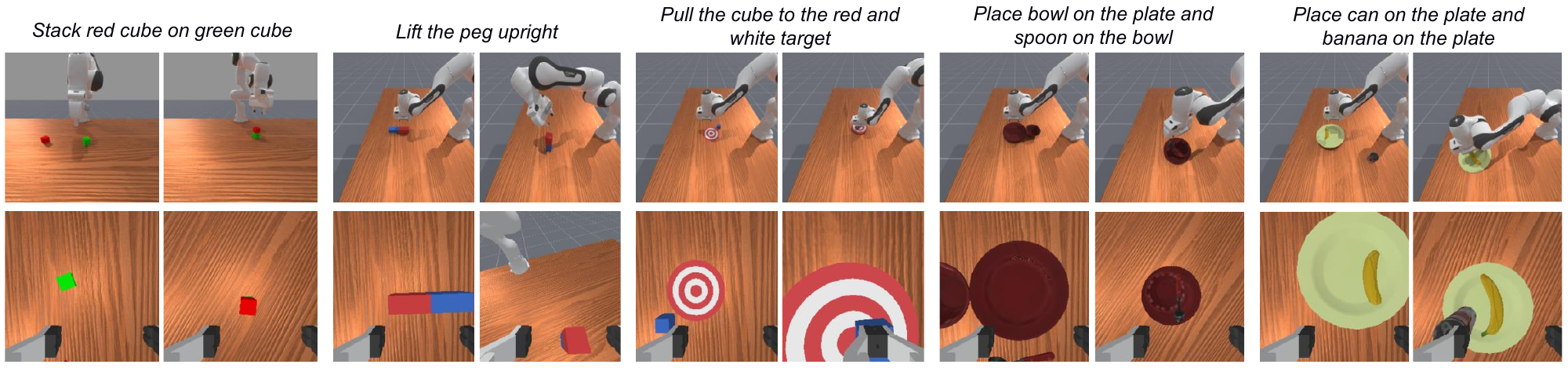}
        \caption{In-domain Environment Setup.}
    \end{subfigure}
    
    \begin{subfigure}[t]{.8\textwidth}
        \centering
        \includegraphics[width=\linewidth]{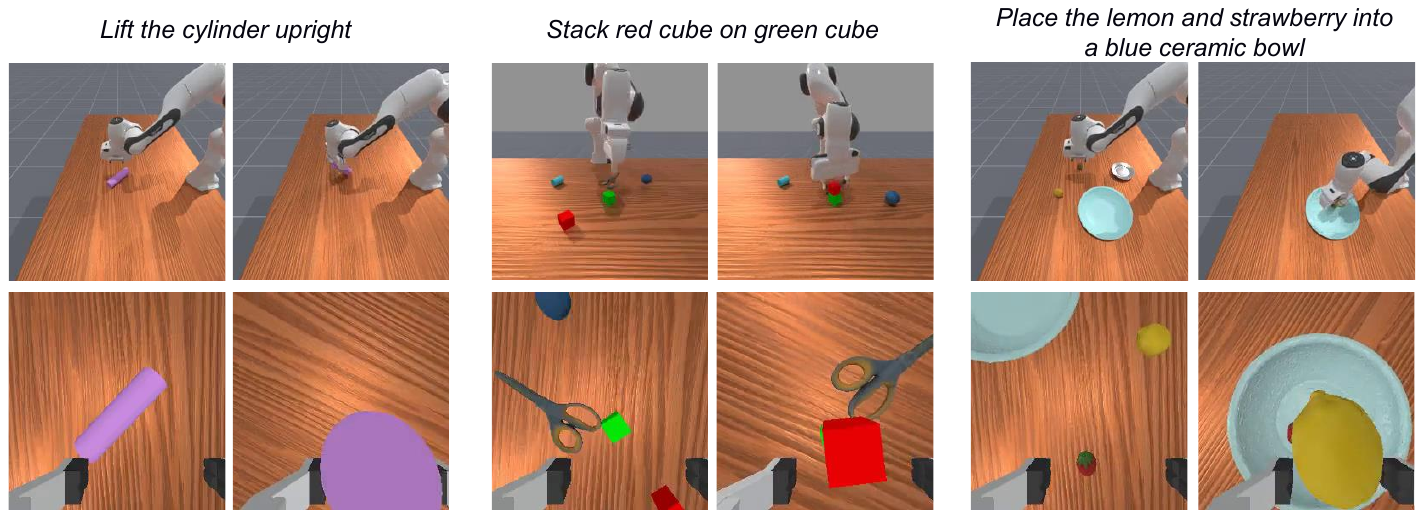}
        \caption{Out-of-domain Environment setup.}
    \end{subfigure}

    \caption{Row 1: exterior camera image, row 2: wrist camera image. For each task, we select one frame at the beginning (left) and end (right) of the trajectories. Corresponding to Figure~\ref{fig:setup_vis} in the main paper}
    \label{fig:setup_vis_suppl}
\end{figure}

More complete environment setup visualizations are presented in Figure~\ref{fig:setup_vis_suppl}.

\section{Justification of the Cosine-Disagreement Guidance Scale}
\paragraph{Adaptive Negative Guidance via Score Disagreement.}

At diffusion timestep $\eta$, let $s_{\textcolor{green}{\textcolor{green}{\text{succ}}}}(\va) = \nabla \log p_s(\va_{\eta})$ and 
$s_{\textcolor{red}{\text{fail}}}(\va_{\eta}) = \nabla \log p_f(\va_{\eta})$ denote the score functions of the success and failure models, respectively. We consider a guided score of the form (\cf Eq. (7) in the main paper):
\begin{equation}
s_\lambda(\va_{\eta}) 
= s_{\textcolor{green}{\text{succ}}}(\va_{\eta}) - \lambda_{\eta}(\va_{\eta})\, s_{\textcolor{red}{\text{fail}}}(\va_{\eta}),
\end{equation}
where $\lambda_{\eta}(\va_{\eta})$ is an adaptive
guidance scale.

\paragraph{Optimal Guidance as Projection.}

Assume there exists an unknown optimal score function $s^*(\va_{\eta})$ that induces task-optimal behavior. We seek $\lambda_{\eta}$ such that $s_\lambda$ best approximates $s^*$ in a least-squares sense:
\begin{equation}
\lambda_{\eta}^*(\va_{\eta})
=
\arg\min_{\lambda}
\left\|
s_{\textcolor{green}{\text{succ}}}(\va_{\eta}) - \lambda\, s_{\textcolor{red}{\text{fail}}}(\va_{\eta}) - s^*(\va_{\eta})
\right\|_2^2.
\end{equation}

Taking derivative with respect to $\lambda$ and setting it to zero yields:
\begin{equation}
\lambda_{\eta}^*(\va_{\eta})
=
\frac{
s_{\textcolor{red}{\text{fail}}}(\va_{\eta})^\top \big(s_{\textcolor{green}{\text{succ}}}(\va_{\eta}) - s^*(\va_{\eta})\big)
}{
\|s_{\textcolor{red}{\text{fail}}}(\va_{\eta})\|_2^2
}.
\end{equation}

\paragraph{Residual Error Approximation.}

We model the optimal score as a first-order correction of the success score:
\begin{equation}
s^*(\va_{\eta}) = s_{\textcolor{green}{\text{succ}}}(\va_{\eta}) + \delta(\va_{\eta}),
\end{equation}
where $\delta(\va_{\eta})$ denotes the residual error of the success model. Substituting gives:
\begin{equation}
\lambda_{\eta}^*(\va_{\eta})
=
- \frac{
s_{\textcolor{red}{\text{fail}}}(\va_{\eta})^\top \delta(\va_{\eta})
}{
\|s_{\textcolor{red}{\text{fail}}}(\va_{\eta})\|_2^2
}.
\end{equation}

Since $\delta(\va_{\eta})$ is not directly observable, we approximate it using the discrepancy between success and failure scores:
\begin{equation}
\delta(\va_{\eta}) \approx s_{\textcolor{green}{\text{succ}}}(\va_{\eta}) - s_{\textcolor{red}{\text{fail}}}(\va_{\eta}).
\end{equation}

Substituting this approximation:
\begin{align}
\lambda_{\eta}^*(\va_{\eta})
&\approx
- \frac{
s_{\textcolor{red}{\text{fail}}}^\top (s_{\textcolor{green}{\text{succ}}} - s_{\textcolor{red}{\text{fail}}})
}{
\|s_{\textcolor{red}{\text{fail}}}\|^2
} \\
&=
1 - \frac{s_{\textcolor{red}{\text{fail}}}^\top s_{\textcolor{green}{\text{succ}}}}{\|s_{\textcolor{red}{\text{fail}}}\|^2}.
\end{align}

\paragraph{Cosine Similarity Approximation.}

Assuming comparable magnitudes $\|s_{\textcolor{green}{\text{succ}}}\| \approx \|s_{\textcolor{red}{\text{fail}}}\|$, we obtain:
\begin{equation}
\lambda_{\eta}^*(\va_{\eta})
\approx
1 - \cos\big(s_{\textcolor{green}{\text{succ}}}(\va_{\eta}),\, s_{\textcolor{red}{\text{fail}}}(\va_{\eta})\big).
\end{equation}

\paragraph{Interpretation.}

This result shows that the optimal guidance scale is proportional to the angular disagreement between the success and failure scores. When the two scores are aligned, the residual correction is small and negative guidance is suppressed. When they disagree, the failure score provides complementary information, and stronger guidance is applied. This leads to a simple and fully local approximation:
\begin{equation}
\lambda_{\eta}(\va_{\eta})
\propto
1 - \cos\big(s_{\textcolor{green}{\text{succ}}}(\va_{\eta}),\, s_{\textcolor{red}{\text{fail}}}(\va_{\eta})\big),
\end{equation}
which avoids explicit estimation of distribution distances while retaining a principled connection to optimal score approximation.

\section{Additional Experimental Results}
\label{sec:additional_exp}
\textbf{Additional Experiments on LIBERO-PRO} We performed additional experiments on LIBERO-PRO \cite{liu2023libero,zhou2025liberopro} and report the success rates of the base $\pi_{0.5}$, using the results reported in the original publication, and our proposed $\text{DAG-VLA}_{\pi_{0.5}}$ with AFIL on LIBERO-10 and LIBERO-Spatial in Table~\ref{tab:libero_perturbations}.
The results show that our proposed $\text{DAG-VLA}_{\pi_{0.5}}$ consistently outperforms the baseline $\pi_{0.5}$ pipeline across five perturbation settings introduced in LIBERO-PRO benchmark -- object attributes (Obj.), initial positions (Pos.), task variations (Task), semantic variations (Sem.), and environmental changes (Env.) -- demonstrating improved generalization through adaptive negative guidance.

\begin{table*}[h!]
\centering
\caption{Success rates (\%) under object (Obj.), position (Pos.), semantic (Sem.), task, and environment (Env.) perturbations on LIBERO-10 and LIBERO-Spatial. Bold values indicate improvements over the corresponding $\pi_{0.5}$ baseline.}
\label{tab:libero_perturbations}
\resizebox{.8\textwidth}{!}{%
\begin{tabular}{llccccc}
\toprule
Method & Benchmark & Obj. & Pos. & Sem. & Task & Env. \\
\midrule
$\pi_{0.5}$ & LIBERO-10
& 92.0 & 0.08 & 93.0 & 0.01 & 46.0 \\
\textbf{DAG-VLA$_{\pi_{0.5}}$ with AFIL} & LIBERO-10
& \textbf{93.4} & \textbf{8.2} & \textbf{93.6} & \textbf{7.0} & \textbf{51.0} \\
\midrule
$\pi_{0.5}$ & LIBERO-Spatial
& 97.0 & 20.0 & 97.0 & 0.01 & 46.0 \\
\textbf{DAG-VLA$_{\pi_{0.5}}$ with AFIL} & LIBERO-Spatial
& \textbf{97.5} & \textbf{46.8} & \textbf{97.4} & \textbf{0.2} & 46.0 \\
\bottomrule
\end{tabular}
}
\end{table*}

\textbf{Ablation Study of Guidance Scale for Static Failure-Informed Learning}
In this section, we analyze the effect of the guidance scale $\lambda$ on the performance of the proposed DAG-VLA with failure-informed learning under the static guidance setting, corresponding to ``DAG-VLA+Static-FIL'' in Table~1 of the main paper. We evaluate on two tasks: \textit{Lift the peg upright} (short-horizon) and \textit{Place bowl on the plate and spoon on the bowl} (long-horizon), in the in-domain setting. The results shown in Table~\ref{tab:abl_lambda} show that $\lambda=0.05$ empirically achieves the best performance across all tasks. In contrast, setting $\lambda$ either too small or too large degrades manipulation performance, indicating the importance of properly balancing attraction toward successful behaviors and repulsion from failure-prone actions.

\begin{table*}[h!]
\centering
\small
\renewcommand{\arraystretch}{1.2}
\caption{Ablation study on guidance scale $\lambda$ for static failure-informed learning. Success rates reported in \%.}
\label{tab:abl_lambda}
\resizebox{.8\textwidth}{!}{%
\begin{tabular}{l|cccc}
\toprule
\textbf{Task} &
$\lambda$ = 0.01 &
$\lambda$ = 0.05 &
$\lambda$ = 0.1 &
$\lambda$ = 0.2 \\
\toprule
\textit{Lift the peg upright} & 84.7 & \textbf{86.7} & 82.7 & 78.0 \\
\textit{Place bowl on the plate and spoon on the bowl} & 46.8 & \textbf{49.3} & 47.1 & 44.0 \\
\bottomrule
\end{tabular}%
}
\end{table*}



\textbf{Error Bars} While diffusion-based policies introduce inherent stochasticity during inference, as they generate actions through an iterative denoising process initialized from random noise. We report standard deviation for proposed DAG-VLA with AFIL in in-domain setting (\cf Table 1 in the main paper) in Table~\ref{tab:std} and Table~\ref{tab:std_ood}. 

\begin{table}[h!]
\centering
\caption{Standard deviation (in percentage points) of 3 test-runs (per task) on DAG-VLA with AFIL in Maniskill~\cite{gu2023maniskill2} in-domain evaluation settings. Each test-run with 50 configs $\times$ 3 runs = 150 rollouts. }
\label{tab:std}
\begin{tabular}{lcc}
\toprule
\textbf{Task} & \textbf{Std Dev ($s$)} \\
\midrule
\textit{Stack red cube on green cube} & <0.01 \\
\textit{Lift the peg upright}   & 2.03 \\
\textit{Pull cube to the red and white target}  & 1.39 \\
\textit{Place bowl on the plate and spoon on the bowl}  & 1.41 \\
\textit{Place can on the plate and banana on the plate}  & 4.44 \\
\bottomrule
\end{tabular}
\end{table}

\begin{table}[h!]
\centering
\caption{Standard deviation (in percentage points) of 3 test-runs (per task) on DAG-VLA with AFIL in Maniskill~\cite{gu2023maniskill2} out-of-domain evaluation settings. Each test-run with 25 configs $\times$ 3 runs = 75 rollouts. }
\label{tab:std_ood}
\begin{tabular}{lcc}
\toprule
\textbf{Task} & \textbf{Std Dev ($s$)} \\
\midrule
\textbf{Lift Cylinder} (unseen object/color) & 1.26 \\
\textbf{Fruit Sorting} (unseen objects + clutter)  & <0.01 \\
\textbf{Stack Cubes} (cluttered scene)  & 3.33 \\
\bottomrule
\end{tabular}
\end{table}

\textbf{Failure Data Efficiency} To further analyze the data efficiency of the failure generator, We therefore conduct a failure-data scaling study using 0\% (corresponding to the “DAG-VLA, no AFIL” baseline in Table 1), 50\%, 75\%, 90\%, and 100\% (corresponding to “DAG-VLA with Adaptive-FIL” in Table 1) of the available raw failure trajectories, while keeping the original training data, model initialization, and evaluation protocol fixed. We evaluate one representative short-horizon task, \textit{Stack red cube on green cube}, and one long-horizon task, \textit{Place can on the plate and banana on the plate}, and report the success rates in Table~\ref{tab:failure_data_scaling}. The results show consistent improvement as more failure trajectories are introduced. Notably, using only 50\% of the available failures already improves the success rate by 4.0\% on the short-horizon task and 12.7\% on the long-horizon task. Increasing the amount of failure data further yields additional but diminishing gains: with 90\% of the data, performance reaches 84.0\% and 70.7\%, close to the full-data results of 84.7\% and 71.3\%, respectively. These results indicate that useful failure-informed guidance can be learned from substantially fewer than 1,500 failure trajectories. At the same time, additional failure data continue to improve performance, suggesting that broader coverage of failure-prone behaviors remains beneficial. 

\begin{table*}[h!]
\centering
\caption{Task success rates (\%) as the number of failure trajectories used for AFIL training increases. The percentage in each column indicates the fraction of the 1,500 available failure trajectories used.}
\label{tab:failure_data_scaling}
\resizebox{.95\textwidth}{!}{%
\begin{tabular}{lccccc}
\toprule
Task
& \shortstack{0 failures\\(0\%)}
& \shortstack{750 failures\\(50\%)}
& \shortstack{1,125 failures\\(75\%)}
& \shortstack{1,350 failures\\(90\%)}
& \shortstack{1,500 failures\\(100\%)} \\
\midrule
\textit{Stack red cube on green cube}
& 76.7 & 80.7 & 82.0 & 84.0 & \textbf{84.7} \\
\textit{Place can on the plate and banana on the plate}
& 56.0 & 68.7 & 69.3 & 70.7 & \textbf{71.3} \\
\bottomrule
\end{tabular}
}
\end{table*}

\textbf{Justification of Dual Action Generator Design} A single action generator conditioned on trajectory quality, as in $\pi_{0.7}$, is a reasonable alternative for learning from mixed-quality data. To evaluate this design, we trained a $\pi_{0.5}$-based single action generator (Single-AG) using outcome-conditioned language prompts: successful trajectories were paired with “success: [task description],” while failed trajectories were paired with “failure: [task description].” However, this formulation consistently underperformed the proposed dual-generator (Dual-AG) design, with both approaches evaluated under the same training and evaluation protocols: 
\begin{table*}[h!]
\centering
\caption{Task success rates (\%) on ManiSkill comparing single- and dual-action-generator VLA models with AFIL. The dual-generator model achieves higher success rates on all five tasks.}
\label{tab:dag_design}
\resizebox{\textwidth}{!}{%
\begin{tabular}{lcc}
\toprule
Task & Single-AG VLA with AFIL & Dual-AG VLA with AFIL \\
\midrule
\textit{Stack red cube on green cube}                   & 72.0 & \textbf{84.7} \\
\textit{Lift the peg upright}                           & 75.3 & \textbf{90.7} \\
\textit{Pull cube to the red and white target}          & 82.5 & \textbf{98.0} \\
\textit{Place bowl on the plate and spoon on the bowl}  & 45.3 & \textbf{56.7} \\
\textit{Place can on the plate and banana on the plate} & 54.7 & \textbf{71.3} \\
\bottomrule
\end{tabular}
}
\end{table*}
We observed similar degradation with alternative prompt formulations, including explicit statements such as “this is a successful/failed trajectory” placed before or after the task description. Therefore, we use separate success and failure generators with disjoint supervision. This design prevents raw failure actions from directly altering the success generator and its parameter statistics, better leverages the visual-language alignment learned during pretraining, and provides explicit estimates of the success and failure action distributions for guidance.

\textbf{Success Only Baseline Ablation} We perform an additional ablation comparing DAG-VLA trained with 750 successful and 750 failure-corrected trajectories against a success-only model trained with 1,500 successful trajectories, corresponding to the same setting as Table 1 in the main paper. The results are reported in Table~\ref{tab:failure_correction_data}.

\begin{table*}[h!]
\centering
\caption{Task success rates (\%) for DAG-VLA trained on 1,500 success-only trajectories or on 750 success and 750 failure-correction trajectories.}
\label{tab:failure_correction_data}
\resizebox{\textwidth}{!}{%
\begin{tabular}{lcc}
\toprule
Task
& \shortstack{DAG-VLA\\(1,500 success-only)}
& \shortstack{DAG-VLA\\(750 success / 750 failure-correction)} \\
\midrule
\textit{Stack red cube on green cube}
& 71.3 & 74.0 \\
\textit{Pull cube to the red and white target}
& 60.0 & 82.5 \\
\textit{Place can on the plate and banana on the plate}
& 39.3 & 46.0 \\
\bottomrule
\end{tabular}
}
\end{table*}

\section{Parameter Analysis and Runtime}
\textbf{Parameter Analysis} For the proposed DAG-VLA built upon $\pi_{0.5}$, we introduce only an additional action generator, implemented as a Gemma-300M expert contributing approximately $428$M parameters, together with a dedicated output projection head containing only $32$K parameters. Since the original $\pi_{0.5}$ model contains approximately $3.35$B parameters, $\text{DAG-VLA}_{\pi_{0.5}}$ introduces a moderate increase in total parameter count, by $12.76$\%, while achieving consistent performance improvements across diverse manipulation tasks, environments, and domains. 

\textbf{Runtime Analysis} We further analyze the inference-time overhead of the proposed adaptive failure-informed guidance. To ensure a fair comparison, we evaluate both the baseline $\pi_{0.5}$ VLA and DAG-VLA with AFIL over 10 episode rollouts under the same environment and hardware setting. 
The average runtime per action horizon step is $2.88$ ms for baseline $\pi_{0.5}$, and $3.89$ ms for DAG-VLA with AFIL, on a NVIDIA A100-SXM4-80GB GPU. The two generators can also be inferred/evaluated in parallel, reducing the runtime to 3.45 ms per action-horizon step, with GPU memory usage increasing only modestly from 14,426.61 MiB to 14,675.53 MiB. 
The moderate runtime increase is expected, as AFIL queries both the success and failure action generators with per-diffusion-step distance measuring during sampling, but remains practical for deployment while providing consistent robustness gains.

\section{Significance Analysis}
we conducted a matched A/B test on one short-horizon task, \textit{Lift the Peg Upright}, and one long-horizon tas, \textit{Place Can on the Plate and Banana on the Plate}, to isolate the effect of adaptive failure guidance. The resuls are shown in Table~\ref{tab:adaptive_fil_paired}. For each task, the treatment and control conditions use the same trained DAG-VLA model, training data, evaluation configurations, and rollout protocol as reported in Table 1 of the main paper. The control condition disables failure-informed guidance and samples actions using only the success generator, corresponding to ``DAG-VLA, no FIL” in Table 1. In contrast, the treatment condition applies Adaptive-FIL during action sampling, corresponding to ``DAG-VLA + Adaptive-FIL” in Table 1. Both conditions are evaluated on the same 50 configurations, with three paired episodes per configuration, yielding 150 paired trials. 

\begin{table*}[h!]
\centering
\caption{Paired comparison of task success rates with and without Adaptive-FIL. Improvements and 95\% paired confidence intervals (CI) are reported in percentage points (pp); $p$-values are from exact McNemar tests.}
\label{tab:adaptive_fil_paired}
\resizebox{\textwidth}{!}{%
\begin{tabular}{lccccc}
\toprule
Task & No AFIL & Adaptive-FIL & Improvement
& 95\% Paired CI & \shortstack{Exact McNemar\\Test ($p$)} \\
\midrule
\textit{Lift the Peg Upright}
& 80.00\% & \textbf{90.67\%} & \textbf{+10.67 pp}
& \textbf{[+3.97, +17.37] pp} & \textbf{0.0037} \\
\textit{Place Can on the Plate and Banana on the Plate}
& 56.00\% & \textbf{71.33\%} & \textbf{+15.33 pp}
& \textbf{[+5.26, +25.41] pp} & \textbf{0.0052} \\
\bottomrule
\end{tabular}
}
\end{table*}

For \textit{Lift the Peg Upright}, Adaptive-FIL succeeds in 136 of 150 trials, compared with 120 of 150 trials without AFIL. Among the discordant pairs, AFIL succeeds in 22 cases in which the control fails, while the control succeeds in only six cases in which AFIL fails. The exact McNemar test yields $p=0.0037$, with a paired odds ratio of 3.67.
For \textit{Place Can on the Plate and Banana on the Plate}, Adaptive-FIL succeeds in 107 of 150 trials, compared with 84 of 150 trials without AFIL. AFIL uniquely succeeds in 43 discordant pairs, compared with 20 control-only successes, yielding an exact McNemar test result of $p=0.0052$ and a paired odds ratio of 2.15.

Because repeated episodes within the same configuration may be correlated, we additionally performed configuration-level analyses. For the two tasks, cluster-bootstrap confidence intervals for the success-rate improvements are $[+3.33, +18.00]$ and $[+4.00, +26.67]$ percentage points, respectively. Configuration-level permutation tests also confirm significant improvements, with $p=0.0119$ and $p=0.0165$, respectively.
After applying the Holm correction across the two primary McNemar tests, both improvements remain statistically significant. These results provide evidence on representative short- and long-horizon tasks that the gains from Adaptive-FIL are unlikely to arise from evaluation variability alone.

\section{Qualitative Failure Analysis}
 To better characterize the failures collected by our pipeline, we conducted a post hoc taxonomy analysis of 150 randomly sampled failed episodes, comprising 424 attempt-level video segments. Qwen3-VL-32B-Instruct was used to generate preliminary annotations, which were then manually reviewed and corrected. Each segment was assigned a primary failure category, an optional secondary category, and a brief visual rationale. Importantly, these annotations are used solely for analysis; AFIL training does not require failure-category labels or taxonomy supervision.

To avoid overcounting repeated attempts within an episode, we define the episode-level root cause as the earliest meaningful failure event. The resulting distribution is shown in Table~\ref{tab:root_failure_modes}. The collected rollouts therefore contain multiple semantically distinct failure modes spanning perception and target selection, reaching and grasping, object transport, release timing, and final placement stability. Representative examples include grasping the wrong object, closing the gripper beside the target, losing the object during transport, releasing it before reaching the target, and producing an initially valid but unstable stack. These failure categories were not specified during data collection or model training; they were identified only through a post hoc analysis of naturally occurring policy rollouts. 

\begin{table}[h!]
\centering
\caption{Distribution of root failure modes across 424 failed episodes. Grasp failure accounts for the largest share.}
\label{tab:root_failure_modes}
\begin{tabular}{lcc}
\toprule
Root Failure Mode & Episodes & Percentage \\
\midrule
Grasp failed             & 267 & 63.0\% \\
No action attempted      &  96 & 22.6\% \\
Wrong-object interaction &  22 &  5.2\% \\
Released too early       &  16 &  3.8\% \\
Misaligned reach         &  12 &  2.8\% \\
Misaligned placement     &   8 &  1.9\% \\
Unstable stack           &   2 &  0.5\% \\
Dropped during transport &   1 &  0.2\% \\
\bottomrule
\end{tabular}
\end{table}

\begin{figure}[h!]
    \centering \includegraphics[width=\linewidth]{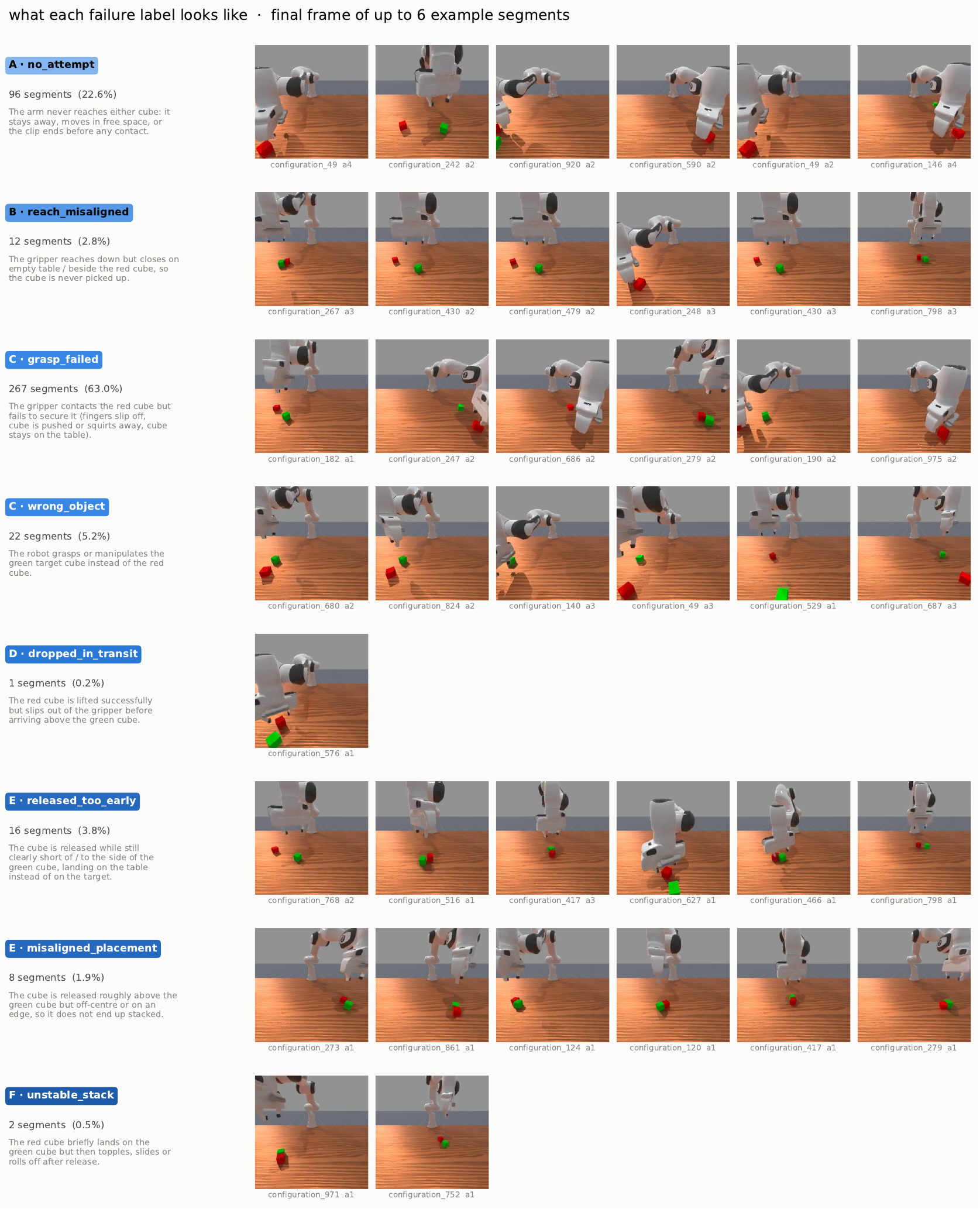}
    \caption{Qualitative analysis of the collected failure rollouts for training our proposed DAG-VLA with adaptive negative guidance (AFIL).}
    \label{fig:label_gallery}
\end{figure}

Although grasp failures constitute the majority of failures for this task, the collected data also capture less frequent but behaviorally distinct errors, including target confusion, transport failure, premature release, placement misalignment, and instability after placement. This long-tailed distribution illustrates why a fixed, manually designed perturbation set may not fully represent the errors encountered by the learned policy. A qualitative of the failure modes is shown in Figure~\ref{fig:label_gallery}.

\section{Limitations and Societal Impact}
\label{sec:limitations}
While AFIL provides an effective mechanism for leveraging failure trajectories in VLA learning, it has several limitations. First, the proposed failure-informed guidance is specifically designed for diffusion- and flow-based generative VLA policies, where action generation is performed through iterative sampling. Extending the same learning-from-failure principle to non-generative or directly autoregressive VLA architectures, such as OpenVLA~\cite{kim2024openvla}-style policies, would require additional adaptation, since these models do not naturally expose intermediate denoising or flow steps for guidance injection. 
Second, although AFIL is model-agnostic within the class of diffusion- and flow-based VLA models, our current experiments are built on the $\pi_{0.5}$ architecture~\cite{black2025pi_}. We leave the evaluation of AFIL on a broader range of generative VLA backbones to future work.

\paragraph{Societal Impact.}
AFIL aims to improve the robustness and reliability of VLA-based robotic manipulation by enabling policies to learn from failures and avoid failure-prone actions during deployment. This may benefit applications such as assistive robotics, household manipulation, warehouse automation, and human-robot collaboration, where more reliable manipulation can reduce task disruption and potential physical damage. However, more robust robotic autonomy also raises safety and accountability concerns. AFIL does not guarantee safe behavior under all distribution shifts, hardware conditions, or human interactions. Real-world deployment should therefore include appropriate safety constraints, monitoring, physical safeguards, and human oversight, particularly in safety-critical environments. Failure data should also be collected and curated responsibly to avoid reinforcing unsafe or domain-specific behaviors. 





\end{document}